\documentclass[10pt,twocolumn,letterpaper]{article}

\usepackage{iccv}

\iccvfinalcopy 

\def\iccvPaperID{22} 

\ificcvfinal\fi

\usepackage[utf8]{inputenc} 
\usepackage[T1]{fontenc}    
\usepackage{hyperref}       
\usepackage{url}            
\usepackage{booktabs}       
\usepackage{amsfonts}       
\usepackage{nicefrac}       
\usepackage{microtype}      
\usepackage{times}
\usepackage{latexsym}
\usepackage{url}
\usepackage{mathtools}
\usepackage{CJKutf8}
\usepackage{array,multirow}
\usepackage{subcaption}
\usepackage{amsmath,amssymb,amsthm}
\usepackage[ruled]{algorithm2e}
\usepackage{comment}
\usepackage{color,soul}
\usepackage{authblk}

\newcommand{\name}{HM-NAS}

\DeclareMathOperator*{\argmin}{arg\,min}
\title{HM-NAS: Efficient Neural Architecture Search via Hierarchical Masking}

\begin{document}
        \vspace{-6mm}
		\author{Shen Yan$^\dagger$, Biyi Fang$^\dagger$, Faen Zhang$^\ddagger$, Yu Zheng$^\dagger$, Xiao Zeng$^\dagger$, Hui Xu$^\ddagger$, Mi Zhang$^\dagger$ \\
		\vspace{-3.5mm}
		$^\dagger$Michigan State University, $^\ddagger$AInnovation \\
		\vspace{1.5mm}
		{\tt\small \{yanshen6, fangbiyi, zhengy30, zengxia6, mizhang\}@msu.edu, \\ 
		\vspace{-0.5mm}
		\tt\small \{zhangfaen, xuhui\}@ainnovation.com} \\
		\vspace{-5mm}
		
	}
\maketitle
    \ificcvfinal\thispagestyle{empty}\fi

\begin{abstract}

The use of automatic methods, often referred to as Neural Architecture Search (NAS), in designing neural network architectures has recently drawn considerable attention.
In this work, we present an efficient NAS approach, named {\name}, that generalizes existing weight sharing based NAS approaches.
Existing weight sharing based NAS approaches still adopt hand designed heuristics to generate architecture candidates.
As a consequence, the space of architecture candidates is constrained in a subset of all possible architectures, making the architecture search results sub-optimal.
%
{\name} addresses this limitation via two innovations.
First, {\name} incorporates a multi-level architecture encoding scheme to enable searching for more flexible network architectures.
Second, it discards the hand designed heuristics and incorporates a hierarchical masking scheme that automatically learns and determines the optimal architecture.
%
Compared to state-of-the-art weight sharing based approaches, {\name} is able to achieve better architecture search performance and competitive model evaluation accuracy.
%
Without the constraint imposed by the hand designed heuristics, our searched networks contain more flexible and meaningful architectures that existing weight sharing based NAS approaches are not able to discover.

\end{abstract}

\section{Introduction}
\label{sec.intro}

Neural architecture search (NAS) has recently attracted significant interests due to its capability of automating neural network architecture design and its success in outperforming hand-crafted architectures in many important tasks such as image classification~\cite{zoph2018nasnet},  object detection~\cite{chen2019detnas}, and semantic segmentation~\cite{nekrasov2019fast}. 
%
%
In early NAS approaches,
architecture candidates are first sampled from the search space; the weights of each candidate are learned independently and are discarded if the performance of the architecture candidate is not competitive~\cite{pham2018enas,zoph2018nasnet,baker2016designing, real2018amoebanet}.
Despite their remarkable performance, since each architecture candidate requires a full training, these approaches are computationally expensive, consuming hundreds or even thousands of GPU days in order to find high-quality architectures.

To overcome this bottleneck, a majority of recent efforts focuses on improving the computation efficiency of NAS using the \textit{weight sharing} strategy~\cite{pham2018enas, liu2018darts, xie2018snas, cai2018proxylessnas, Brock2018SMASHOM}. %
Specifically, 
%
%
rather than training each architecture candidate independently, the architecture search space is encoded within a single over-parameterized \textit{supernet} which includes all the possible connections (i.e., wiring patterns) and operations (\emph{e.g.}, convolution, pooling, identity).
The supernet is trained \textit{only once}.
All the architecture candidates inherit their weights directly from the supernet without training from scratch.
%
By doing this, the computation cost of NAS is significantly reduced. 
%

Unfortunately, although the supernet subsumes all the possible architecture candidates, existing weight sharing based NAS approaches still adopt \textit{hand designed heuristics} to extract architecture candidates from the supernet.
%
As an example, in many existing weight sharing based NAS approaches such as DARTS~\cite{liu2018darts}, the supernet
is organized as stacked cells and each cell contains multiple nodes connected with edges.
However, when extracting architecture candidates from the supernet, each candidate is hard coded to have \textit{exactly} two input edges for each node with \textit{equal importance} and to associate each edge with \textit{exactly} one operation.
%
As such, the space of architecture candidates is constrained in a subset of all possible architectures, making the architecture search results sub-optimal.

\begin{figure}[t]
\centering
\includegraphics[scale=0.43]{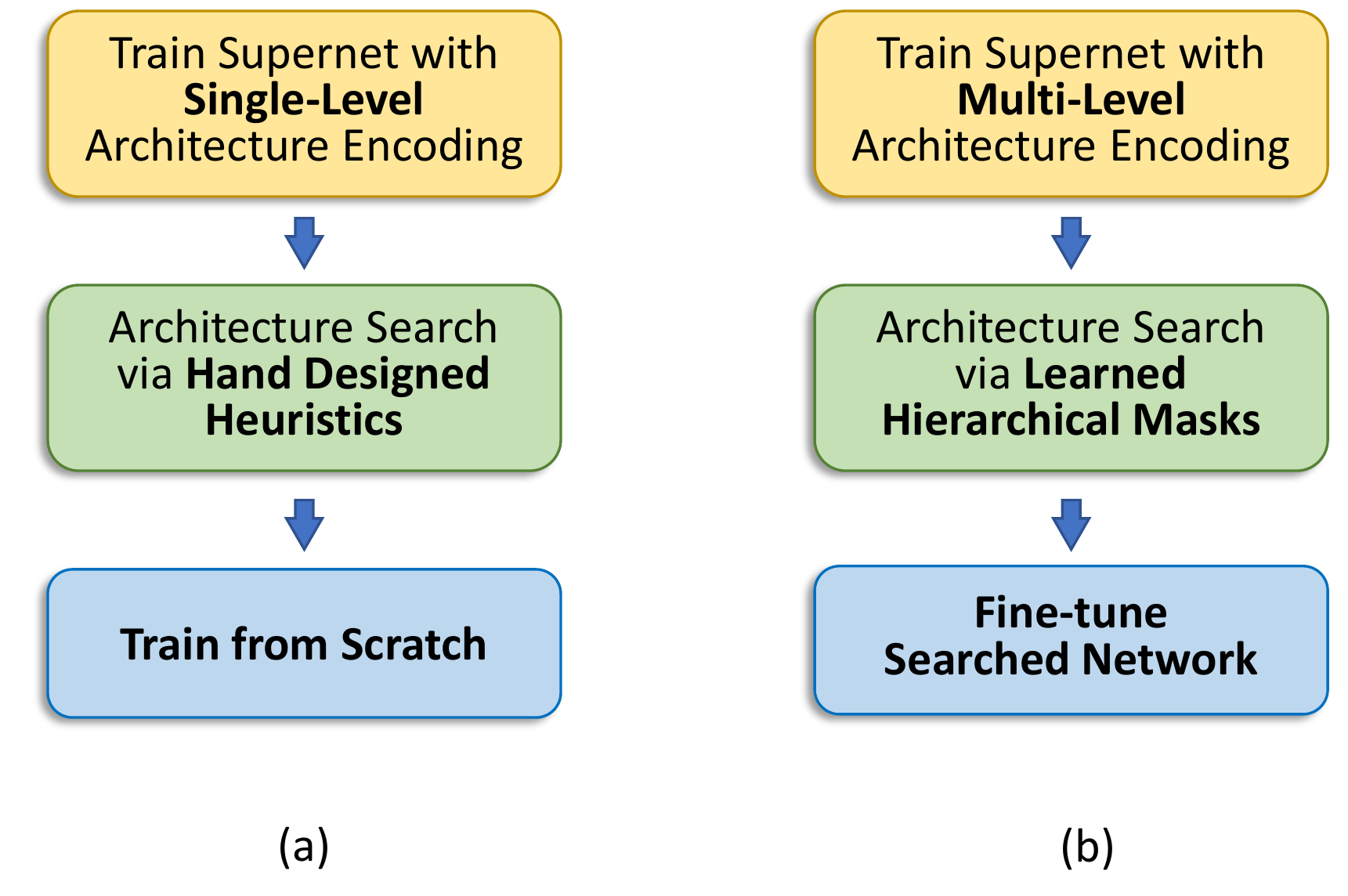}
\vspace{-1mm}
\caption{The pipelines of (a) existing weight sharing based NAS approaches such as DARTS~\cite{liu2018darts} and SNAS \cite{xie2018snas}; and (b) {\name} (our approach).
}
\label{dia.maskfig1}
\vspace{-2mm}
\end{figure}

%
%
Given the constraint of existing weight sharing approaches, it is natural to ask the question:
\textit{will we be able to improve architecture search performance if we loosen this constraint?} 
To this end, we present {\name}, an efficient neural architecture search approach that effectively addresses such limitation of existing weight sharing based NAS approaches to achieve better architecture search performance and competitive model evaluation accuracy.
As illustrated in Figure \ref{dia.maskfig1}, to loosen the constraint, {\name} incorporates a \textit{multi-level architecture encoding} scheme which enables an architecture candidate extracted from the supernet to have arbitrary numbers of edges and operations associated with each edge.
Moreover, it allows each operation and edge to have different weights which reflect their relative importance across the entire network.
%
Based on the multi-level encoded architecture, {\name} formulates neural architecture search as a model pruning problem: it discards the hand designed heuristics and employs a \textit{hierarchical masking} scheme to automatically learn the optimal numbers of edges and operations and their corresponding importance as well as mask out unimportant network weights.
Moreover, the addition of these learned hierarchical masks on top of the supernet also provides a mechanism to help  correct the architecture search bias caused by bilevel optimization of architecture parameters and network weights during supernet training~\cite{sciuto2019evaluating, bender2018oneshot, guo2019single}.
%
Because of such benefit, {\name} is able to use the unmasked network weights to speed up the training process.
We evaluate {\name} on both CIFAR-10 and ImageNet and our results are promising:
{\name} is able to achieve competitive accuracy on CIFAR-10 with $1.6\times$ to $1.8\times$ less parameters and $2.7\times$ total training time speed-up compared with state-of-the-art weight sharing approaches. 
%
Similar results are also achieved on ImageNet.
%
Moreover, we have conducted a series of ablation studies that demonstrate the superiority of our multi-level architecture encoding and hierarchical masking schemes over randomly searched architectures, as well as single-level architecture encoding and hand designed heuristics used in existing weight sharing based NAS approaches.
Finally, we have conducted an in-depth analysis on the best-performing network architectures found by {\name}.
Our results show that without the constraint imposed by the hand designed heuristics, our searched networks contain more flexible and meaningful architectures that existing weight sharing based NAS approaches are not able to discover. 

\vspace{1mm}  
In summary, our work makes the following contributions:
%
\begin{itemize}

  \vspace{-1mm}    
  \item We present {\name}, an efficient neural architecture search approach that loosens the constraint of existing weight sharing based NAS approaches.
  
  \vspace{-1.5mm} 
  \item We introduce a multi-level architecture encoding scheme which enables an architecture candidate to have arbitrary numbers of edges and operations with different importance.
  We also introduce a hierarchical masking scheme which is able to not only automatically learn the optimal numbers of edges, operations and important network weights, but also help correct the architecture search bias  caused by bilevel optimization during supernet training.

  
  \vspace{-1.5mm} 
  \item Extensive experiments show that compared to state-of-the-art weight sharing based NAS approaches, {\name} is able to achieve better architecture search efficiency and competitive model evaluation accuracy.
  
  
  
  
\end{itemize}



\section{Related Work}
\label{sec.related}


Designing high-quality neural networks requires domain knowledge and extensive experiences. To cut the labor intensity, there has been a growing interest in developing automated neural network design approaches through NAS.
Pioneer works on NAS employ reinforcement learning (RL) or evolutionary algorithms to find the best architecture based on nested optimization~\cite{pham2018enas,zoph2018nasnet,baker2016designing, real2018amoebanet}. 
However, these approaches are incredibly expensive in terms of computation cost.
For example, in \cite{zoph2018nasnet}, it takes 450 GPUs for four days to search for the best network architecture.

To reduce computation cost, many works adopt the weight sharing strategy where the weights of architecture candidates are inherited from a supernet that subsumes all the possible architecture candidates.
%
To further reduce the computation cost, recent weight sharing based approaches such as DARTS \cite{liu2018darts} and SNAS \cite{xie2018snas} replace the discrete architecture search space with a continuous one and employ gradient descent to find the optimal architecture.
However, these approaches restrict the continuous search space with hand designed heuristics, which could jeopardize the architecture search performance.  
Moreover, as discussed in \cite{sciuto2019evaluating, bender2018oneshot, guo2019single}, the  bilevel optimization of architecture parameters and network weights used in existing weight sharing based approaches inevitably introduces bias to the architecture search process, making their architecture search results sub-optimal.
%
Our approach is related to DARTS and SNAS in the sense that we both build upon the weight sharing strategy.
However, our goal is to address the above limitations of existing approaches to achieve better architecture search performance.



\begin{table}[t] 
\centering
\scalebox{0.8}{
\begin{tabular}{lccc}
\hline

\multicolumn{1}{l}{\textbf{NAS Approach}} & \multicolumn{1}{c}{\textbf{\begin{tabular}[c]{@{}c@{}}Architecture \\ Encoding\end{tabular}}} & \multicolumn{1}{c}{\textbf{\begin{tabular}[c]{@{}c@{}}Retrain \\ from Scratch\end{tabular}}} & \multicolumn{1}{c}{\textbf{\begin{tabular}[c]{@{}c@{}}Use \\ Proxy\end{tabular}}} \\

\hline
ENAS~\cite{pham2018enas} & Operations  & Yes & Yes  \\
NASNet~\cite{zoph2018nasnet}  & Operations  & Yes & Yes    \\
AmoebaNet~\cite{real2018amoebanet} & Operations  & Yes & Yes    \\
NAONet~\cite{luo2018naonet} & Operations  & Yes  & Yes  \\
ProxylessNAS~\cite{cai2018proxylessnas} & Operations  & Yes & No   \\
FBNet~\cite{fbnet} & Operations  & Yes & No  \\ 
\hline
DARTS \cite{liu2018darts} & Operations  & Yes & Yes   \\
SNAS \cite{xie2018snas} & Operations  & Yes & Yes  \\
\hline
\textbf{{\name}} & \textbf{Operations \& Edges}  & \textbf{No} &  \textbf{No} \\
\hline
\end{tabular}
}
\vspace{-1mm}
\caption{Comparison between HM-NAS and other NAS approaches on a number of important dimensions.}
\vspace{-2mm}
\label{tab.overview}
\end{table}

Our approach is also related to ProxylessNAS~\cite{cai2018proxylessnas}.
ProxylessNAS formulates NAS as a model pruning problem.
In our approach, the employed hierarchical masking scheme also prunes the redundant parts of the supernet to generate the optimal network architecture.
The distinction is that ProxylessNAS focuses on pruning operations (referred to as path in \cite{cai2018proxylessnas}) of the supernet, while {\name} provides a more generalized model pruning mechanism which prunes the redundant operations, edges, and network weights of the supernet to derive the optimal architecture.
Our approach is also similar to ProxylessNAS as being a proxyless approach. 
Rather than adopting a proxy strategy like~\cite{liu2018darts, xie2018snas}, which transfers the searched architecture to another larger network, both {\name} and ProxylessNAS directly search the architectures on target datasets without architecture transfer.
%
However, unlike ProxylessNAS which involves retraining as the last step, {\name} eliminates the prolonged retraining process and replaces it with a fine-tuning process with the reuse of the unmasked pretrained supernet weights.

Table \ref{tab.overview} provides a comparison between {\name} and relevant approaches on a number of important dimensions. 
The combination of the proposed multi-level architecture encoding and hierarchical masking techniques makes {\name} superior over many existing approaches.
We quantify such superiority in \S\ref{sec.evaluation}.

\label{sec.related}
 
\section{Our Approach}

\subsection{Search Space and Supernet Design with Multi-Level Architecture Encoding}
\label{sec.supernet}

Following~\cite{real2018amoebanet,liu2018darts,xie2018snas}, 
we use a cell structure with an ordered sequence of nodes as our search space.
The network is then composed of several identical cells which are stacked on top of each other.
%
Specifically, a cell is represented using a directed acyclic graph (DAG) where each node $x$ in the DAG is a latent representation (\emph{e.g.}, a feature map in a convolutional network).
A cell is set to have two input nodes, one or more intermediate nodes, and one output node.
Specifically, the two input nodes are connected to the outputs of cells from two previous cells; each intermediate node is connected by all its predecessors; and the output node is the concatenation of all the intermediate nodes within the cell.



To build the supernet that subsumes all the possible architectures in the search space, existing works such as DARTS~\cite{liu2018darts} and SNAS~\cite{xie2018snas} associate each edge in the DAG with a mixture of candidate operations (\emph{e.g.}, convolution, pooling, identity) instead of a definite one.
Moreover, each candidate operation of the mixture is assigned with a learnable variable (\textit{i.e.}, operation mixing weight) which encodes the importance of this candidate operation. 
As such, the mixture of candidate operations associated with a particular edge is represented as the softmax over all candidate operations:
\begin{equation}
    \overline{o}(x) = \sum_{i = 1}^{N} \frac{\text{exp}(\alpha_i)}{\sum_{j} \text{exp}(\alpha_j)} o_i(x)
\end{equation}
where $\{o_i\}$ denote the set of $N$ candidate operations, $\{\alpha_i\}$ denote the set of $N$ real-valued operation mixing weights. 

 

Although this supernet encodes the importance of different candidate operations within each edge, \textit{it does not provide a mechanism to encode the importance of different edges across the entire DAG}.
Instead, all the edges across the DAG are constrained to have the same importance.
%
However, as we observed in our experiments (\S\ref{sec.evaluation}), loosening this constraint is able to help NAS find better architectures.
%




%
Motivated by this observation, in {\name}'s supernet, besides encoding the importance of each candidate operation within an edge, we introduce a separate set of learnable variables (\textit{i.e.}, edge mixing weights) to \textit{independently encode the importance of each edge across the DAG}.
%
%
As such, each intermediate node $x^{(i)}$ in the DAG is computed based on all of its predecessors as: 
\begin{equation}
    x^{(i)} = \sum_{j < i} \frac{\text{exp}(\beta^{(i,j)})}{\sum_{k < i} \text{exp}(\beta^{(i,k)})} \overline{o}(x^{(j)})
\end{equation}
where $\beta^{(i,j)}$ denote the real-valued edge mixing weight for the directed edge $(i,j)$.

In summary, $\boldsymbol{\alpha}=\{\alpha_i\}$ encode the architecture at the \textit{operation level} while $\boldsymbol{\beta}=\{\beta^{(i,j)}\}$ encode the architecture at the \textit{edge level}. 
Therefore, we have constructed a supernet with \textit{multi-level architecture encoding} where $\boldsymbol{\alpha}$ and $\boldsymbol{\beta}$ altogether encode the overall architecture of the network,
and we refer to $\{\boldsymbol{\alpha},\boldsymbol{\beta}\}$ as the \textit{architecture parameters}.




\begin{figure*}[t]
\centering
\includegraphics[scale=0.65]{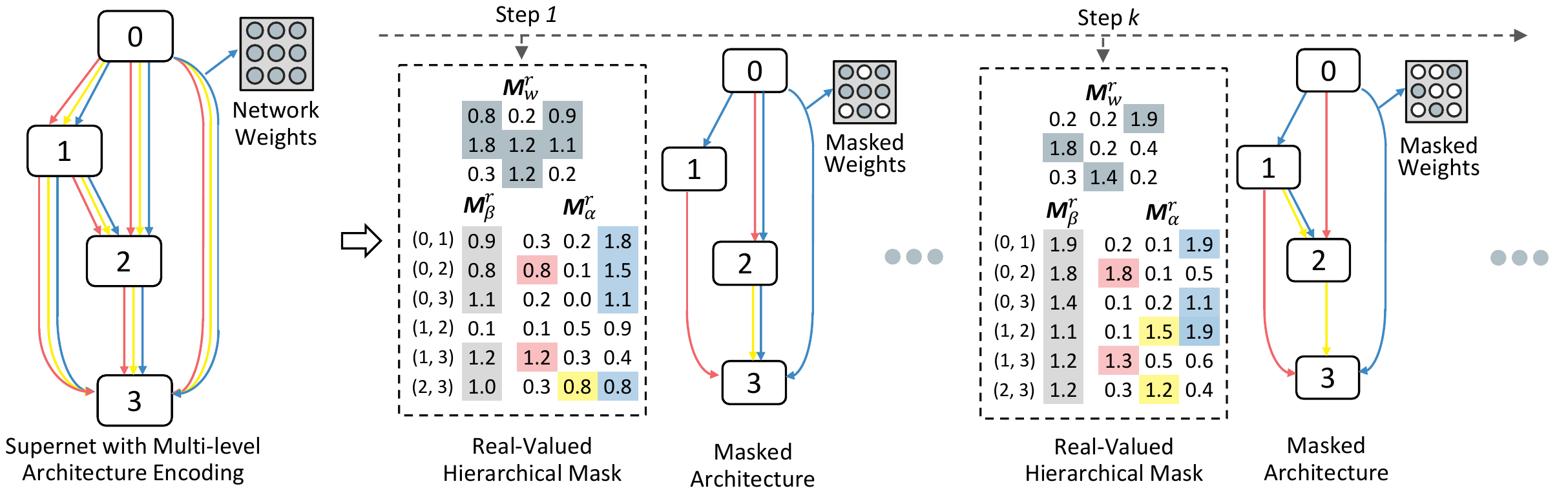}

\caption{Illustration of the iterative hierarchical masking process on a single cell. In this example, each edge has 3 candidate operations marked using red, yellow, and blue color respectively. 
In each iteration, the real-valued hierarchical masks $\{\boldsymbol{M}^{r}_{\alpha}, \boldsymbol{M}^{r}_{\beta}, \boldsymbol{M}^{r}_{w}\}$ are passed through a deterministic thresholding function to obtain the corresponding binary masks (highlighted grids represent `1', the rest represent `0') that mask out redundant operations, edges, and weights of the supernet.}
\label{dia.mask}
\end{figure*}

\subsection{Training the Supernet}
\label{sec.train}

To train the multi-level encoded supernet, we follow \cite{liu2018darts} to jointly optimize the architecture parameters $\{\boldsymbol{\alpha},\boldsymbol{\beta}\}$ and the network weights $\boldsymbol{w}$ in a bilevel way via stochastic gradient descent with first or second-order approximation. 
Let $\mathcal{L}_{train}$ and $\mathcal{L}_{val}$ denote the training loss and validation loss respectively.
The goal is to find $\{\boldsymbol{\alpha}^*,\boldsymbol{\beta}^*\}$ that minimize $\mathcal{L}_{val}(\boldsymbol{\alpha}, \boldsymbol{\beta}, \boldsymbol{w}^*)$, where $\boldsymbol{w}^*$ is obtained by minimizing the training loss $\boldsymbol{w}^* = \argmin_{w} \mathcal{L}_{train}(\boldsymbol{\alpha}^*, \boldsymbol{\beta}^*, \boldsymbol{w})$.
For the details of this bilevel optimization, please refer to \cite{liu2018darts} as we do not claim any new contribution on this part.


Here we want to emphasize two techniques that we find helpful in training our multi-level encoded supernet.
First, due to insufficient training of network weights $\boldsymbol{w}$ at the beginning of the supernet training, the architecture parameters $\boldsymbol{\alpha}$ and $\boldsymbol{\beta}$ could be randomly selected. To avoid this, similar to \cite{fbnet}, we adopt a warm start in training of $\boldsymbol{w}$ while freezing the training of $\boldsymbol{\alpha}$ and $\boldsymbol{\beta}$. 
Second, updating $\boldsymbol{\alpha}$ and $\boldsymbol{\beta}$ too frequently could lead to underfitting of $\boldsymbol{w}$. We solve this by triggering the optimization of $\boldsymbol{\alpha}$ and $\boldsymbol{\beta}$ stochastically rather than doing it constantly, with a probability of $p = \sigma(\textit{iter})$, where $\textit{iter}$ is the number of iterations and $\sigma(\cdot)$ is a monotonically non-increasing function that satisfies $\sigma(0)=1$. After a prolonged decrease, the probability $p$ may even be set to zero, \textit{i.e.}, no bilevel optimization is conducted any longer and only $\boldsymbol{w}$ is optimized.
%

%

\subsection{Searching the Optimal Architecture via \\ Hierarchical Masking}
\label{sec.masking}

Given the trained supernet, we formulate neural architecture search as a model pruning problem, and iteratively prune the redundant \textit{operations}, \textit{edges}, and \textit{network weights} of the supernet in a hierarchical manner to derive the optimal architecture through a scheme which we refer to as \textit{hierarchical masking}.

%
Figure~\ref{dia.mask} illustrates the iterative hierarchical masking process on a single cell.
Specifically, we begin with the trained supernet as our base network, and initiate three types of real-valued masks for operations, edges, and network weights, respectively.
These masks are passed through a deterministic thresholding function to obtain the corresponding binary masks.
%
These generated binary masks are then elementwisely multiplied with the architecture parameters $\{\boldsymbol{\alpha}^*, \boldsymbol{\beta}^*\}$ and network weights $\boldsymbol{w}^*$ of the supernet to generate a searched network. 
By iteratively training the real-valued masks through backpropagation combined with network binarization techniques~\cite{courbariaux2015binaryconnect} in an end-to-end manner, the binary masks learned in the end are able to mask out redundant operations, edges, and network weights in the supernet to derive the optimal architecture.

%
%

Formally, let $\boldsymbol{M}^r=\{\boldsymbol{M}^{r}_{\alpha}, \boldsymbol{M}^{r}_{\beta}, \boldsymbol{M}^{r}_{w}\}$ denote the real-valued hierarchical masks, where $\boldsymbol{M}^{r}_{\alpha}, \boldsymbol{M}^{r}_{\beta}, \boldsymbol{M}^{r}_{w}$ is the real-valued mask for operations, edges, and network weights, respectively.
Architecture search is then reduced to finding $\boldsymbol{M}^{r*}$ which minimizes the training loss of the masked supernet:
\vspace{-2mm}
\begin{equation}
\label{eq.mask}
    \boldsymbol{M}^{r*} = \argmin_{\boldsymbol{M}^{r}}\mathcal{L}(\mathcal{P}_{\boldsymbol{M}}(\boldsymbol{\alpha}^*, \boldsymbol{\beta}^*,\boldsymbol{w}^*))
\vspace{-3mm}
\end{equation}

\begin{equation}
\label{eq.threshold}
    \boldsymbol{M} = H(\boldsymbol{M}^{r} - \tau)
\end{equation}
where $\boldsymbol{M} =\{\boldsymbol{M}_{\alpha}, \boldsymbol{M}_{\beta}, \boldsymbol{M}_{w}\}$ are the corresponding binary masks, $H(\cdot)$ is the Heaviside step function as the deterministic thresholding function, $\tau$ is the pre-defined threshold, and $\mathcal{P}(\cdot)$ is the elementwise projection function. In this work, we use elementwise multiplication for $\mathcal{P}(\cdot)$. 

Even though the Heaviside step function in (\ref{eq.threshold}) is non-differentiable, we adopt the approximation strategy used in BinaryConnect~\cite{courbariaux2015binaryconnect} to approximate the gradients of real-valued masks $\boldsymbol{M^r}$ using the gradients of the binary masks $\boldsymbol{M}$, and thus update the real-valued masks $\boldsymbol{M^r}$ using the gradients of the binary masks $\boldsymbol{M}$. 
As shown in prior works~\cite{mallya2018piggyback, courbariaux2015binaryconnect,hubara2016binarized}, this strategy is effective because the gradients of $\boldsymbol{M}$ actually act as a regularizer or a noisy estimator of the gradients of $\boldsymbol{M^r}$.
By doing this, the binary masks can be trained in an end-to-end differentiable manner.

%

%
%

\subsection{Deriving the Final Model via Fine-Tuning}
\label{sec.finetune}
The hierarchical masking process in \S\ref{sec.masking} outputs not only the optimal network architecture but also a set of optimized network weights.
As such, we can derive the final model via fine-tuning instead of retraining the searched architecture from scratch.
%
%
With the optimized network weights, the searched architecture is able to maintain comparable accuracy compared to the supernet (\emph{e.g.}, $\small{\sim}1\%$ loss on CIFAR-10), and thus acts as a significantly better starting point for fine-tuning.
This not only ensures higher accuracy, but also replaces the prolonged retraining process with a more efficient fine-tuning process, as we will demonstrate in \S\ref{sec.evalcifar}.


\begin{algorithm}[t]
\vspace{1mm}
\small
    \textbf{Input:} multi-level architecture encoded supernet $\boldsymbol{\Theta}(\boldsymbol{\alpha}, \boldsymbol{\beta}, \boldsymbol{w})$, real-valued masks $\boldsymbol{M}^{r}_{\alpha}$, $\boldsymbol{M}^{r}_{\beta}$,  $\boldsymbol{M}^{r}_{w}$,  threshold $\tau$, deterministic thresholding function $H(\cdot)$, elementwise projection function $\mathcal{P}(\cdot)$
    
    \textbf{Output:} \big\{$\widehat{\boldsymbol{\Theta}}(\boldsymbol{
    \alpha^{*}}, \boldsymbol{\beta^{*}}, \boldsymbol{w^{*}}), \boldsymbol{M}^{*}_{\alpha}, \boldsymbol{M}^{*}_{\beta}, \boldsymbol{M}^{*}_{w} $\big\}: the optimized searched model and binary masks
    
    \tcp{supernet training}\label{supernet training}
    Initialize $\boldsymbol{\alpha}
    \leftarrow \boldsymbol{\alpha^{0}}$, $\boldsymbol{\beta} \leftarrow \boldsymbol{\beta^{0}}$, $\boldsymbol{w} \leftarrow \boldsymbol{w^{0}}$, $t \leftarrow 0$.

    \While{not converge}{
        Update $\boldsymbol{\beta}$ and $\boldsymbol{\alpha}$ by descending $\nabla_{\boldsymbol{\beta}}\mathcal{L}_{val}(\boldsymbol{\alpha^{*}, \beta, w^{*}})$ and $\nabla_{\boldsymbol{\alpha}}\mathcal{L}_{val}(\boldsymbol{\alpha, \beta^{*}, w^{*}})$ with a probability of $\sigma(t)$ 
        
        Update $\boldsymbol{w}$ by descending $\nabla_{\boldsymbol{w}} \mathcal{L}_{train}(\boldsymbol{\alpha^{*}, \beta^{*}, w})$ 
        
        $t \leftarrow t + 1$
       
    }
    
    \tcp{searching via hierarchical masking} \label{masking}
    Initialize  $\boldsymbol{M}^{r}_{\alpha} \leftarrow \boldsymbol{M}^{0}_{\alpha}$, $\boldsymbol{M}^{r}_{\beta} \leftarrow \boldsymbol{M}^{0}_{\beta}$, $\boldsymbol{M}^{r}_{w} \leftarrow \boldsymbol{M}^{0}_{w}$
    
    \While{not converge}{
        Feed forward and loss calculation with
        $\mathcal{P}_{H(\boldsymbol{M}^{r}_{w} - \tau)}(\boldsymbol{w^{*}})$,
        $\mathcal{P}_{H(\boldsymbol{M}^{r}_{\alpha} - \tau)}(\boldsymbol{\alpha}^{*})$, $\mathcal{P}_{H(\boldsymbol{M}^{r}_{\beta} - \tau)}(\boldsymbol{\beta}^{*})$
        
        Update $\boldsymbol{M}^r$ by descending $\nabla_{H(\boldsymbol{M}^r - \tau)}
        \mathcal{L}_{train}(\mathcal{P}_{{\boldsymbol{M}}}(\boldsymbol{\alpha}^{*}, \boldsymbol{\beta}^{*},
        \boldsymbol{w^{*}}))$
        
    }
    \tcp{fine-tuning the searched network} \label{fine-tuning}
    
    
    Initialize $\boldsymbol{w} \leftarrow \boldsymbol{w^{*}}$. Construct searched network $\widehat{\boldsymbol{\Theta}}(\boldsymbol{\alpha}^{*}, \boldsymbol{\beta}^{*}, \boldsymbol{w})$ masked by $\boldsymbol{M}^{*}_{\alpha}, \boldsymbol{M}^{*}_{\beta}, \boldsymbol{M}^{*}_{w}$. 
    
    \While{not converge}{
        Update unmasked $\boldsymbol{w}$ by descending $\nabla_{\boldsymbol{w}} \mathcal{L}_{train}(\mathcal{P}_{{\boldsymbol{M}}^{*}}(\boldsymbol{\alpha}^{*}, \boldsymbol{\beta}^{*},
        \boldsymbol{w}))$
        
    }
    \caption{{\name}}
    \label{alg.1}
\end{algorithm}
\vspace{-1mm}
\label{sec.algo}

\section{Experiments and Results}
\label{sec.evaluation}
%
%

We evaluate the performance of {\name} and compare it with state-of-the-arts NAS approaches on two benchmark datasets: CIFAR-10 (\S\ref{sec.evalcifar}) and ImageNet (\S\ref{sec.evalimagenet}).  
%
Moreover, we have conducted a series of ablation studies that validate the importance and effectiveness of the proposed multi-level architecture encoding scheme and hierarchical masking scheme incorporated in the design of {\name} (\S\ref{arch.ablation}).
Finally, we provide an in-depth analysis on the architecture found by {\name} (\S\ref{arch.anaysis}).

%


\subsection{Experimental Setup}

We use 3 cells and 36 initial channels to build the supernet for CIFAR-10, and 5 cells and 24 initial channels for ImageNet.
Following DARTS~\cite{liu2018darts}, our cell consists of 7 nodes in all the experiments. The input nodes, \textit{i.e.}, the first and second nodes of cell $k$ is the output of cell $k - 1$ and $k - 2$, respectively. The output node is the depthwise concatenation of all the intermediate nodes.
We include the following operations: $3\times3$ and $5\times5$ separable convolutions, $3\times3$ and $5\times5$ dilated separable convolutions, $3\times3$ max pooling and average pooling, and identity.  ReLU-Conv-BN triplet is adopted for convolutional operations except the first convolutional layer (Conv-BN), and each separable convolution is applied twice. The default stride is 1 for all operations unless the output size is changed. The experiments are conducted using a single NVIDIA Tesla V100 GPU.

\vspace{2mm}
\subsection{Results on CIFAR-10}
\label{sec.evalcifar}

\vspace{0mm}
\noindent
\textbf{Training Details}. \label{cifar10-training-details} 
%
%
We begin with training the supernet for 100 epochs with batch size 128. In each epoch, we first train weights $\boldsymbol{w}$ on $80\%$ of the training set using SGD with momentum. The initial learning rate is 0.1 with decay following a cosine decaying schedule. The momentum is 0.9 and weight decay is 3e-4. Architecture parameters $\boldsymbol{\alpha}$, $\boldsymbol{\beta}$ are randomly initialized and scaled by 1e-3. Next, We train $\boldsymbol{\alpha}$ and $\boldsymbol{\beta}$ on the rest $20\%$ of the training set with Adam optimizer \cite{kingma2014adam} with the learning rate of 3e-4 and weight decay of 1e-3. 
We empirically observe more stable training process when using Adam for optimizing the architecture parameters, which is also used in \cite{liu2018darts}. Following \cite{fbnet}, we postpone the training of $\boldsymbol{\alpha}$ and $\boldsymbol{\beta}$ by 10 epochs to warm up $\boldsymbol{w}$ first. The supernet training takes 7.5 hours (or 30 hours for the second-order approximation). Once the supernet is trained, we perform 20 epochs of neural architecture search via hierarchical masking using the entire training set. Hierarchical masks $\boldsymbol{M}^{r}_{\alpha}, \boldsymbol{M}^{r}_{\beta}, \boldsymbol{M}^{r}_{w}$ are initialized as 1e-2. They are trained using the Adam optimizer with an initial learning rate of 1e-4 for $\boldsymbol{M}^{r}_{w}$ and 1e-5 for $\boldsymbol{M}^{r}_{\alpha}$ and $\boldsymbol{M}^{r}_{\beta}$, which is decayed by a factor of 10 after 10 epochs. The binarizer threshold $\tau$ (Equation \ref{eq.threshold}) is 5e-3\footnote{Our method is robust to thresholds in the range of [0,  1e-2].}. The hierarchical mask training takes 3.5 hours. Lastly, the masked network is fine-tuned for 200 epochs for 9.6 hours with cutout~\cite{devries2017cutout}.

\vspace{1mm}
\noindent
\textbf{Architecture Evaluation}. 
Table~\ref{tab.cifar} shows our evaluation results on CIFAR-10 where `c/o' denotes cutout adapted from~\cite{devries2017cutout}. The test error of {\name} is on par with state-of-the-art NAS methods. Notably, {\name} achieves this by using the \textit{fewest} parameters among all methods. Specifically, {\name} only uses 1.8M parameters, which is $1.4\times$ to $3.2\times$ fewer compared to others.

\begin{table*}[h]
\centering
\scalebox{0.86}{
\begin{tabular}{lcccccl} \hline
\multicolumn{1}{c}{\textbf{Architecture}} &   \textbf{\begin{tabular}[c]{@{}c@{}}Test Error \\ (\%)\end{tabular}} & \textbf{\begin{tabular}[c]{@{}c@{}}Params \\ (M)\end{tabular}} & \textbf{\begin{tabular}[c]{@{}c@{}}Search Cost \\ (GPU days)\end{tabular}} & \textbf{\begin{tabular}[c]{@{}c@{}}Train Cost \\ (GPU days)\end{tabular}} & \textbf{\begin{tabular}[c]{@{}c@{}}Total Cost \\ (GPU days)\end{tabular}} &  \textbf{\begin{tabular}[c]{@{}l@{}}Search \\ Method\end{tabular}} \\ \hline
DenseNet-BC~\cite{huang2017densely} & 3.46 & 25.6 & - & - & - & manual \\ \hline
ENAS + c/o~\cite{pham2018enas} & 2.89 & 4.6 & \textbf{0.45} & (630 epochs)$^\dagger$ & - & RL \\
NASNet-A + c/o~\cite{zoph2018nasnet} & 2.65 & 3.3 & 3150 & - & - & RL \\
SNAS + c/o~\cite{xie2018snas} & 2.85 & 2.8 & 1.5 & 1.5 (600 epochs) & 3 & gradient-based \\
ProxyLess-G + c/o & \textbf{2.08} & 5.7 & 4* & (600 epochs) & - & gradient-based \\
AmoebaNet-A + c/o~\cite{real2018amoebanet} & 3.34 $\pm$ 0.06 & 3.2 & 3150 & - & - & evolution \\
AmoebaNet-B + c/o~\cite{real2018amoebanet} & 2.55 $\pm$ 0.05 & 2.8 & 3150 & - & - & evolution \\
DARTS (1st order) + c/o~\cite{liu2018darts} & 3.00 $\pm$ 0.14 & 3.3 & 1.5* & 2$^\dagger$ (600 epochs) & 3.5 & gradient-based \\
DARTS (2nd order) + c/o~\cite{liu2018darts} & 2.76 $\pm$ 0.09 & 3.3 & 4* & 2$^\dagger$ (600 epochs) & 6 & gradient-based \\ \hline
\hline
\textbf{{\name}} (1st order) + c/o & 2.78 $\pm$ 0.07 & \textbf{1.8} & \textbf{0.45} & \textbf{0.4} (200 epochs) & \textbf{0.85} & gradient-based \\ 
\textbf{{\name}} (2nd order) + c/o & 2.41 $\pm$ 0.05 & \textbf{1.8} & 1.4 & \textbf{0.4} (200 epochs) & \textbf{1.8} & gradient-based \\ \hline
\multicolumn{7}{l}{*~Results obtained from authors' official response in openreview.} \\
\multicolumn{7}{l}{$^\dagger$~Results obtained using code publicly released by the authors.}\\
\end{tabular}
}
\vspace{-2mm}
\caption{Comparison with state-of-the-arts on CIFAR-10.}
\vspace{-2mm}
\label{tab.cifar}
\end{table*}

\vspace{1mm}
\noindent
\textbf{Performance at Different Training Stages.}
Table \ref{tab.process} breaks down the complete architecture search process of {\name} and shows the performance of {\name} at different stages. 
Specifically, compared to the supernet, the searched network (derived after hierarchical masking) loses only $\small{\sim}1\%$ accuracy with $40\%$ less parameters. Although directly using this searched network is not optimal (with test error $5.14\%$), it does provide a good initialization for fine-tuning, which leads to lower test error (from $5.14\%$ to $2.41\%$). 
\begin{table}[h] 
\centering
\scalebox{0.83}{
\begin{tabular}{cccc}
\hline
\textbf{Architecture} & \textbf{\begin{tabular}[c]{@{}c@{}}Test \\ Error\\ (\%)\end{tabular}} & \textbf{\begin{tabular}[c]{@{}c@{}}Params\\ (M)\end{tabular}} & \textbf{\begin{tabular}[c]{@{}c@{}}Params\\ Reduction\\ (\%)\end{tabular}} \\ \hline
Supernet & 4.2 & 3 & - \\
Searched Network & 5.14 & 1.8 & 40 \\
Searched Network + Fine-Tuning & \textbf{2.41} & \textbf{1.8} & \textbf{40} \\ \hline
\end{tabular}
}
\vspace{-2mm}
\caption{Performance of {\name} at different architecture search stages.}
\vspace{-3mm}
\label{tab.process}
\end{table}

\begin{figure}[h]
\vspace{-4mm}
\includegraphics[scale=0.46]{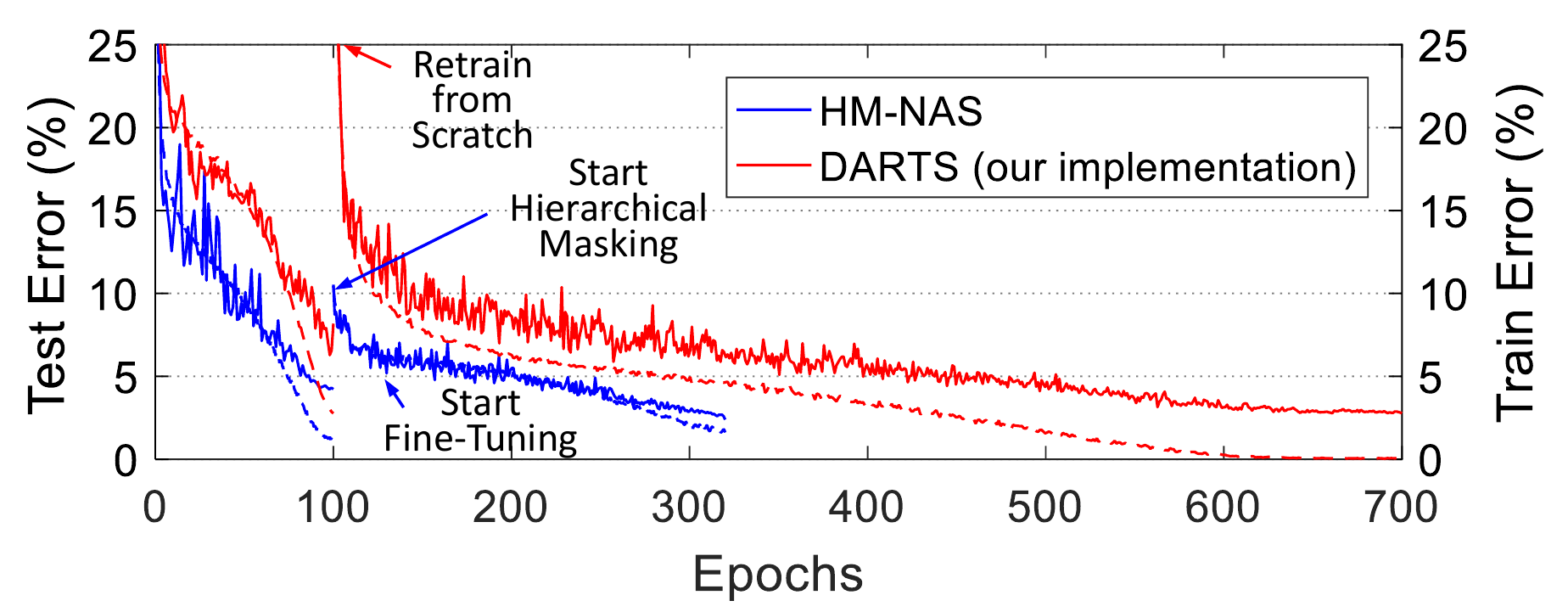}
\vspace{-6mm}
\caption{Training curves of {\name} (in blue color) and 
DARTS (in red color) on CIFAR-10. Solid lines denote test errors (y-axis on the left); dashed lines denote training errors (y-axis on the right).}
\label{dia.cost}
\vspace{-4mm}
\end{figure}

\vspace{0mm}
\noindent
\textbf{Architecture Search Cost Analysis.}
To find the optimal architecture, {\name} only uses 0.85 or 1.8 GPU days, which is significantly faster compared to all other NAS methods. 
To understand why {\name} is efficient, 
we compare the complete architecture search process of {\name} to DARTS. Figure~\ref{dia.cost} illustrates the training curve of {\name} (in blue color) and DARTS\footnote{Our implementation based on the code released by the authors.} (in red color) during the complete architecture search process on CIFAR-10.
Specifically, both {\name} and DARTS use the first 100 epochs to train the supernet with the same train/validation dataset split. 
Due to multi-level architecture encoding, {\name} is able to achieve better test results after 100 epochs. Then, DARTS transfers the learned cell to build a larger network and retrains it from scratch. This process takes approximately 600 epochs to converge.
%
In contrast, from 100 epoch to 120 epoch and onward, {\name} performs architecture search via hierarchical masking and fine-tuning, respectively. This process only takes 220 epochs to converge, which is $2.7\times$ faster compared to DARTS.

\vspace{-2mm}

\vspace{2mm}
\subsection{Results on ImageNet}
\label{sec.evalimagenet}
We conduct experiments on ImageNet 1000-class \cite{imagenet_cvpr09} classification task, where input image size is $224\times224$. The dataset has around 1.28M training images and we test on the 50k validation images. 

\vspace{1mm}
\noindent
\textbf{Training Details}.
We adopt the small computation regime (\emph{e.g.}, MobileNet-V1 \cite{howard2017mobilenetv1}) in the experiments. 
Following \cite{fbnet}, 100 classes from the original 1,000 classes of ImageNet is randomly sampled to train the supernet for 100 epochs with batch size 128. It takes around one GPU day to finish the supernet training. Once the supernet is trained, the hierarchical masking is then performed with the same optimization settings mentioned in \S\ref{cifar10-training-details}. The hierarchical masking process takes around one GPU day to finish. Lastly, the searched network is fine-tuned on the entire ImageNet training dataset (with 1,000 classes) for $60$ epochs with initial learning rate 1e-2 then decreased to 1e-3 at the epoch 30. This phase takes around $3$ GPU days to finish.

\vspace{1mm}
\noindent
\textbf{Architecture Evaluation}. 
Table~\ref{tab.imagenet} shows our evaluation results on ImageNet. The result is comparable to DARTS, considering that we adopt the exact same search space in DARTS~\cite{liu2018darts}, which uses the operations incorporated in MobileNet-V1~\cite{howard2017mobilenetv1}. Notably, we achieve comparable results to the state-of-the-art gradient-based NAS approaches~\cite{xie2018snas, liu2018darts} with 1.2$\times$ to 1.3$\times$ less parameters and 1.08$\times$ to 1.2$\times$ less FLOPs.
With a larger supernet and better candidate operations such as the ones used in  MobileNet-V2~\cite{sandler2018mobilenetv2}, We believe that the results could be further improved.

\begin{table}[t]
\centering
\scalebox{0.87}{
\begin{tabular}{lccc}
\hline
\multicolumn{1}{c}{\textbf{Architecture}} & \textbf{\begin{tabular}[c]{@{}c@{}}Top-1 Acc.\\ (\%)\end{tabular}} & \textbf{\begin{tabular}[c]{@{}c@{}}Params \\ (M)\end{tabular}} & \textbf{\begin{tabular}[c]{@{}c@{}}FLOPs\\ (M)\end{tabular}} \\ \hline
MobileNet-V1~\cite{howard2017mobilenetv1} & 70.6 & 4.2 & 569 \\
MobileNet-V2~\cite{sandler2018mobilenetv2} & \textbf{74.7} & 6.9 & 585 \\
\hline
NASNet-A~\cite{zoph2018nasnet} & 74.0 & 5.3 & 564 \\
Amoeba-A~\cite{real2018amoebanet} & 74.5 & 5.1 & 555 \\
DARTS~\cite{liu2018darts} & 73.3 & 4.7 & 574 \\
SNAS~\cite{xie2018snas} & 72.7 & 4.3 & 522 \\
\hline
\textbf{{\name}} & 73.4 & \textbf{3.6} & \textbf{482} \\ \hline
\end{tabular}
}
\vspace{-0mm}
\caption{Comparison with state-of-the-arts on ImageNet.}
\vspace{-3mm}
\label{tab.imagenet}
\end{table}



\subsection{Ablation Studies}
\label{arch.ablation}
In this section, we conduct a series of ablation studies to demonstrate the superiority of the design of HM-NAS.
The ablation studies are conducted on CIFAR-10 with second-order derivative introduced in Table \ref{tab.cifar}. 

\vspace{1mm}
\noindent
\textbf{Comparison to Single-Level Architecture Encoding.}
To demonstrate the superiority of the proposed multi-level architecture encoding scheme over single-level architecture encoding, we compare the single-level encoded network against the multi-level encoded network, both with hand designed heuristics (by replacing each mixed operation with the most likely operation and taking the top-2 confident edges from distinct nodes).
As shown in Table \ref{tab.cifar.multilevel.ablation}, the multi-level architecture encoding achieves $2.7$\% test error, giving $0.4\%$ accuracy improvement over the single-level one.

\vspace{1mm}
\noindent
\textbf{Comparison to Hand Designed Heuristics.}
To demonstrate the superiority of learned hierarchical masks over hand designed heuristics, 
we compare the multi-level encoded network with learned hierarchical masks against the one with hand designed heuristics.
%
As shown in Table \ref{tab.cifar.multilevel.ablation}, the hierarchical masks achieve $2.41$\% test error, providing about $0.3\%$ accuracy improvement over hand designed heuristics. 

\vspace{1mm}
\noindent
\textbf{Comparison to Random Architectures.}
As discussed in~\cite{sciuto2019evaluating, li2019random}, random architecture is also a competitive choice.
Therefore, we perform a \textit{random architecture search} from the same supernet for 18 times. 
As shown in Table \ref{tab.cifar.ablation}, the average test error of random architecture is $3.41\%$.
This is competitive to the test error of single-level encoded network ($3.1\%$ in Table \ref{tab.cifar.multilevel.ablation}), whose search space is constrained by hand designed heuristics.
Similar findings are also observed in~\cite{sciuto2019evaluating}.
In contrast, {\name} outperforms the random architecture by $1\%$ in test error with 3$\times$ less training epochs.
This is because with multi-level architecture encoding and hierarchical masking, the search space is significantly enlarged, making it challenging for random search to find a competitive network.

\vspace{1mm}
\noindent
\textbf{Comparison to Random Initialization.}
As our final ablation study, to demonstrate the superiority of unmasked network weights obtained from hierarchical masking over random weights, we \textit{randomly initialize} the weights of the searched network (same architecture as {\name}) and train it for $600$ epochs on par with the training setup in DARTS. We run 5 times of random initialization, each running the same number of epochs. As shown in Table \ref{tab.cifar.ablation}, the average test error of random initialization is 2.95\%, which is comparable to DARTS but considerably higher than {\name}. 
%
This result indicates that a good initialization for the searched network is critical for obtaining the best-performing results and fast convergence.

\begin{table}[t]
\centering
\scalebox{0.8}{
\begin{tabular}{cccc}
\hline
\textbf{\begin{tabular}[c]{@{}c@{}}Architecture \\ Encoding\end{tabular}}                                     & \textbf{Derived Rule}      & \textbf{\begin{tabular}[c]{@{}c@{}}Test \\ Error\\ (\%)\end{tabular}} & \textbf{\begin{tabular}[c]{@{}c@{}}Params\\ (M)\end{tabular}} \\ \hline
Single-Level ($\boldsymbol{\alpha}$)                                                                            & Hand Designed Heuristics   & 3.1                                                                   & 2.5                                                           \\ 
\hline
\begin{tabular}[c]{@{}c@{}}

Multi-Level ($\boldsymbol{\alpha}$, $\boldsymbol{\beta}$) \end{tabular} & Hand Designed Heuristics   & 2.7                                                                   & 2.1                                                           \\ \hline
\begin{tabular}[c]{@{}c@{}}

Multi-Level ($\boldsymbol{\alpha}$, $\boldsymbol{\beta}$)  \end{tabular} & Learned Hierarchical Masks & \textbf{2.41}                                        & \textbf{1.8}                                 \\ \hline
\end{tabular}
}
\vspace{-2mm}
\caption{Comparison to single-level architecture encoding and hand designed heuristics.}
\vspace{1mm}
\label{tab.cifar.multilevel.ablation}
\end{table}

\vspace{1mm}
\begin{table}[t]
\centering
\scalebox{0.80}{
\begin{tabular}{cccc} \hline
\multicolumn{1}{c}{\textbf{Architecture}} &   \textbf{\begin{tabular}[c]{@{}c@{}}Test Error \\ (\%)\end{tabular}} & \textbf{\begin{tabular}[c]{@{}c@{}}Params \\ (M)\end{tabular}}  & \textbf{\begin{tabular}[c]{@{}c@{}}Train Cost \\ (epochs) \end{tabular}}  \\ \hline
\hspace{-3mm} Random Architecture & 3.41 $\pm$ 0.15 & 2.1  & 600   \\
Random Initialization $^\ddagger$ & 2.95 $\pm$ 0.08 & 1.8  & 600   \\
\textbf{{\name}}  & \textbf{2.41} $\pm$ 0.05 & \textbf{1.8}  & \textbf{200}  \\ \hline
\multicolumn{4}{l}{$^\ddagger$~Same architecture as {\name} + c/o with random initialized weights.} 
\end{tabular}
}
\vspace{-2mm}
\caption{Comparison to random architectures and random initialization.}
\vspace{-2mm}
\label{tab.cifar.ablation}
\end{table}


\vspace{-0mm}
\begin{figure*}[t]
\centering
\includegraphics[scale=0.51]{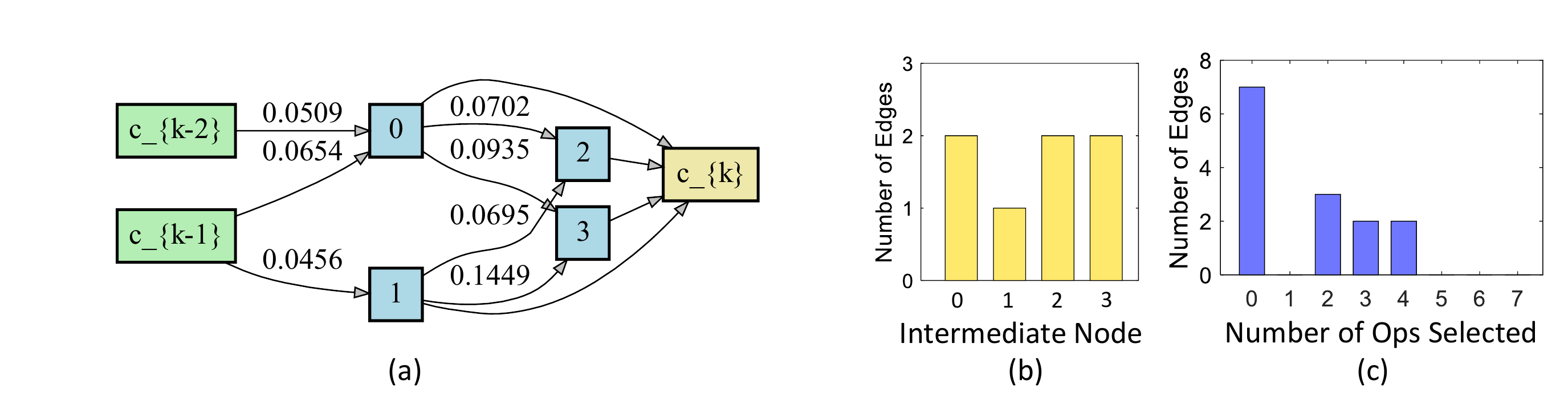}
\vspace{-4mm}
\caption{Details of learned cell for CIFAR-10. (a) cell structure. (b) number of input edges of four intermediate nodes. (c) histogram of the number of edges w.r.t the number of operations selected.}
\label{dia.histo}
\vspace{-2mm}
\end{figure*}

\begin{figure*}[t]
\hspace{-5mm}
\centering
\includegraphics[scale=0.52]{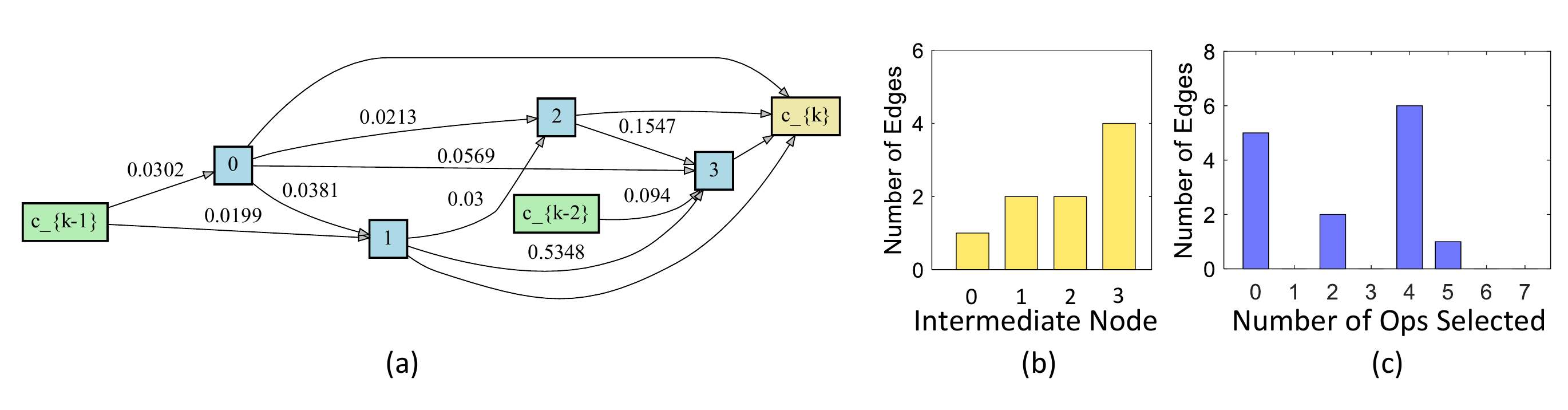}
\vspace{-4mm}
\caption{Details of learned cell for ImageNet. (a) cell structure. (b) number of input edges of four intermediate nodes. (c) histogram of the number of edges w.r.t the number of operations selected.}
\label{dia.imghisto}
\vspace{-3mm}
\end{figure*}

\subsection{Searched Architecture Analysis}
\label{arch.anaysis}

Finally, we provide an in-depth analysis on the network architecture found by {\name}. We have the following three important observations.
\vspace{1mm}

\noindent
\textbf{Different Learned Importance for Different Edges}.
Figure~\ref{dia.histo}(a) and Figure~\ref{dia.imghisto}(a) illustrate the details of the learned cell for CIFAR-10 and ImageNet respectively, where the importance of edge, \textit{i.e.}, edge mixing weight $\beta^{(i, j)}$, is marked above every edge. 
Unlike DARTS in which each edge has the \textit{same hard-coded} importance, due to multi-level architecture encoding, the best-performing cell found by {\name} has \textit{different learned} importance for different edges across the cell.
Moreover, we find that edges connecting to later intermediate nodes have higher importance than early intermediate nodes. One possible explanation is that during cell construction, each intermediate node is ordered and is derived from its predecessors by accumulating information passed from its predecessors. Hence, it has more influence on the output of the cell, which is reflected by the higher importance learned through our approach. 
\textit{Once the importance of the edge is no longer heuristically determined but automatically learned, the multi-level architecture encoding provides a more flexible way to encode the entire supernet architecture and thus provides us with a better superset for architecture search.}

\begin{figure}[h]
\hspace{-3mm}
\includegraphics[scale=0.447]{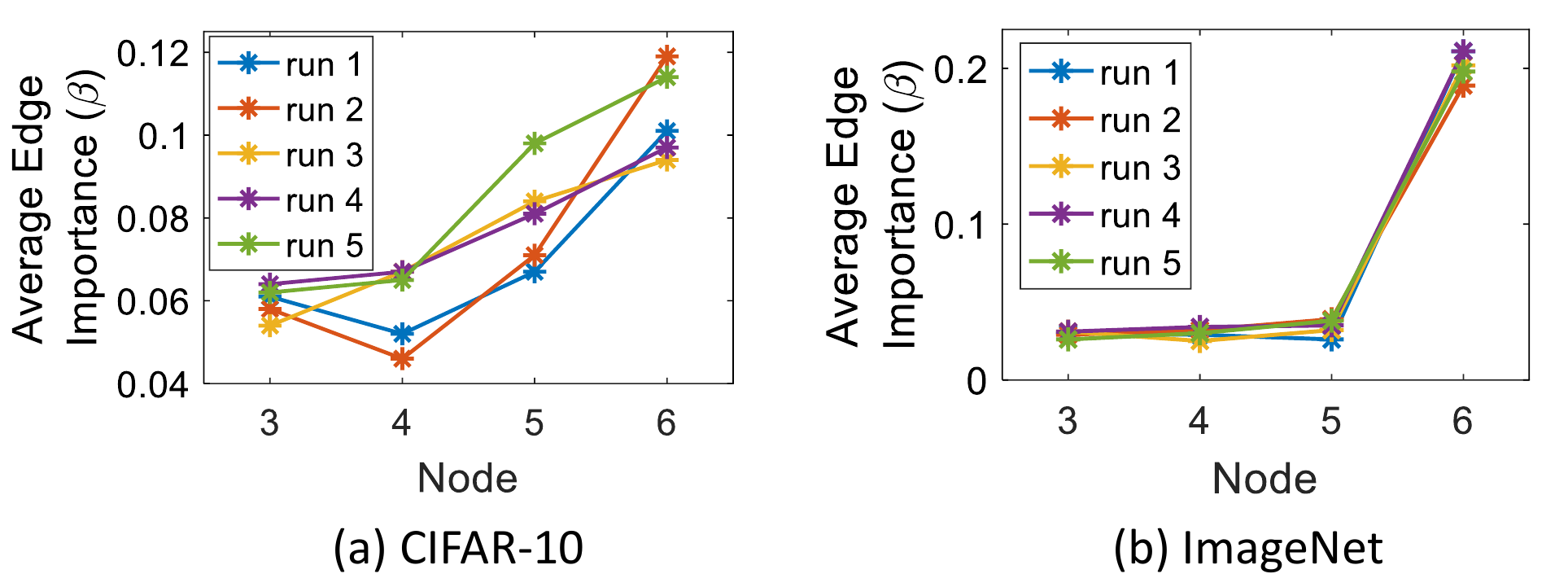}
\vspace{-5mm}
\caption{Robustness of learned edge importance of (a) CIFAR-10 and (b) ImageNet. The learned importance of different edges does not strongly depend on initialization.}
\label{dia.run5}
\vspace{-4mm}
\end{figure}

\vspace{1mm}
\noindent
\textbf{Robustness of Learned Edge Importance}.
%
We repeat the experiments 5 times with random seeds on both CIFAR-10 and ImageNet datasets, and report the (per run) averaged incoming edge importance in each immediate node with the best validation performance of the architecture over epochs (we keep track of the most recent architectures). As shown in Figure~\ref{dia.run5}, we observe that the learned importance of different edges does not strongly depend on initialization: \textit{even if the initial weights are randomly initialized, after the search process completes, the later intermediate nodes always have higher importance than earlier nodes}.     

\vspace{1mm}
\noindent
\textbf{More Flexible Architectures}.
%
Figure~\ref{dia.histo}(b) and Figure~\ref{dia.imghisto}(b) show the number of input edges connecting to each intermediate node, while Figure~\ref{dia.histo}(c) and Figure~\ref{dia.imghisto}(c) show the histogram of the number of edges w.r.t the number of operations selected (\emph{e.g.} the third bar from the left shows that three edges have two associated operations).
Unlike DARTS in which each intermediate node is hard coded to have \textit{exactly} two input edges and each edge is hard coded to have \textit{exactly} one operation, the best-performing cell found by {\name} has intermediate nodes which have more ($\geq$ 2) incoming edges, and edges are associated with zero (the edge is removed) or multiple ($\geq$ 1) operations.
This observation suggests that \textit{{\name} is able to find more flexible architectures that existing weight sharing based NAS approaches are not able to discover}. 

In principle, other constraints such as \textit{the number of cells}, \textit{the number of channels}, \textit{the number of nodes in a cell}, and \textit{the combination operation} (\emph{e.g.} sum, concatenation) can all be further relaxed by the proposed multi-level architecture encoding and hierarchical masking schemes. 
We leave these explorations as our future work.

%




\section{Conclusion}
\label{sec.conc}

We present an efficient NAS approach named {\name} that generalizes existing weight sharing based NAS approaches.
{\name} incorporates a multi-level architecture encoding scheme to enable an architecture candidate to have arbitrary numbers of edges and operations with different importance. 
The learned hierarchical masks not only select the optimal numbers of edges, operations and important network weights, but also help correct the architecture search bias caused by bilevel optimization in supernet training.
%
%
Experiment results show that, compared to state-of-the-arts, {\name} is able to achieve competitive accuracy on CIFAR-10 and ImageNet with improved architecture search efficiency.

\section{Acknowledgement}
\label{sec.ack}
This work was partially supported by NSF
Awards CNS-1617627 and PFI:BIC-1632051.

\bibliographystyle{unsrt}
\bibliography{refs}







\end{document}